\documentclass{article}

\usepackage{PRIMEarxiv}

\usepackage[utf8]{inputenc} % allow utf-8 input
\usepackage[T1]{fontenc}    % use 8-bit T1 fonts
\usepackage{hyperref}       % hyperlinks
\usepackage{url}            % simple URL typesetting
\usepackage{booktabs}       % professional-quality tables
\usepackage{amsfonts}       % blackboard math symbols
\usepackage{nicefrac}       % compact symbols for 1/2, etc.
\usepackage{microtype}      % microtypography
\usepackage{lipsum}
\usepackage{fancyhdr}       % header
\usepackage{times}
\usepackage{amsmath}
\usepackage{amssymb}
\newcommand{\x}{\mathbf{x}}
\newcommand{\g}{\mathbf{g}}
\newcommand{\velocity}{\mathbf{v}}
\newcommand{\avector}{\mathbf{a}}
\newcommand{\bvector}{\mathbf{b}}
\newcommand{\addt}{\ensuremath{\sum_{t=1}^T}}
\newcommand{\stxt}{\ensuremath{s_t^{\eta}(\x_t^{\eta})}}
\newcommand{\stxs}{\ensuremath{s_t^{\eta}(\x^*)}}
\newcommand{\grads}{\ensuremath{\nabla s_t^{\eta}(\x^{\eta})}}
\newcommand{\xte}{\ensuremath{\x_t^{\eta}}}
\usepackage[lined,ruled,commentsnumbered]{algorithm2e}
\usepackage{algpseudocode}

\title{Parameter-free version of Adaptive Gradient Methods for Strongly-Convex Functions
}

\author{
  Deepak Gouda$^*$, Hassan Naveed$^*$, Salil Kamath$^*$\\
  Georgia Institute of Technology \\
  Atlanta, USA\\
  \texttt{\{deepakgouda, hnaveed3, skamath36\}@gatech.edu} \\
}

\begin{document}
\maketitle

\def\thefootnote{*}\footnotetext{These authors contributed equally to this work}

\begin{abstract}
The optimal learning rate for adaptive gradient methods applied to $\lambda$-strongly convex functions relies on the parameters $\lambda$ and learning rate $\eta$. In this paper, we adapt a universal algorithm along the lines of Metagrad, to get rid of this dependence on $\lambda$ and $\eta$. The main idea is to concurrently run multiple experts and combine their predictions to a ``master" algorithm. This master enjoys $O(d\log T)$ regret bounds.
\end{abstract}

% keywords can be removed

\section{Introduction}

Online convex optimization (OCO) is a widely recognized approach for representing sequential decision-making processes. The OCO algorithm operates in the following manner:
\begin{itemize}
    \item the learner selects an action $\x_t$
    \item an adversary chooses a loss function $f_t(\cdot)$
    \item the learner incurs a loss $f_t(\x_t)$
    \item the learner updates its action based on this loss 
\end{itemize} 
The aim is to minimize regret, which is defined as the disparity between the cumulative loss suffered by the learner and that of the optimal action in retrospect \cite{hazan2021introduction}. This is defined as:
$$Regret = \sum_{t = 1}^T f_t(\x_t) - \min_{\x\in D} \sum_{t = 1}^T f_t(\x)$$
Several work has been done on this under various loss function settings. Among the most frequent are convex functions, exponentially concave functions and strongly convex functions. These leverage the slope of the function to alter the gradient update step in order to obtain sublinear regret bounds. OGD on convex functions achieves $O(\sqrt{T})$ regret with step size decreasing by $1/\sqrt{T}$ \cite{OGD_regret}, and a $O(\log(T))$ regret bound on strongly convex functions with a step-size of order $1/T$ \cite{SC_OGD_regret}. Similarly, for exponentially concave functions, the ONS algorithm achieves a regret bound of $O(\log(T))$ as well \cite{SC_OGD_regret}.

Our proposed algorithm combines a variant of the meta algorithm described in \cite{maler_guanghui} using strongly convex experts from a version of Adagrad described in \cite{Adagrad_duchi}.

\section{Related Work}

Recent work on adaptive gradients set a parameter specific learning rate to help with sparse gradients, i.e, settings where we only have a few non-zero features. A popular method was Adagrad, which decays the learning rate faster for parameters with larger gradients \cite{Adagrad_duchi}. This way, we have a larger step size for weights that have smaller magnitude. It still has a regret bound of the order ${O}(\sqrt{T})$ similar to work by \cite{hazan2021introduction}, but this data dependent bound demonstrates better performance in practice. 

The Adagrad algorithm without modification did not guarantee a ${O}(\log (T))$ bound in strongly convex functions. This extension to the strongly convex case was made by \cite{SC_adagrad} by one major change. First, they decayed the step size on the parameters using the sum of squared gradients, rather than taking an L2 norm of them. They also had a decay rate for the constant added  by Adagrad numerical stability, but this did not result in any better bounds or performance.

At another front, there has been a push to create universal algorithms that can perform well for most function types. As previously mentioned, the type of alogrithm and learning rate strictly divided the type of functions into three main categories: convex, exp-concave, and strongly convex. It would be preferable to have algorithms that work for not only these three cases at once, but also everything in between. This influenced work on an algorithm called \textit{MetaGrad} by \cite{metagrad}, which works for functions that satisfy the so called Bernstien condition. It consists of one master algorithm which spawns several experts. Each of these slaves has its own learning rate and employs a surrogate loss function which closely mimics the Online Newton Step (ONS). The master alogrithm combines the predictions from each of these slaves and passes down its gradients to them after incurring a loss. This approach only requires gradient information and achieves  $O(\sqrt{T \log T \log T})$ regret on convex and $O(d\log T)$ on exponentially concave functions.

However, the $O(d\log T)$ regret Metagrad implies on strongly convex functions has an $O(d)$ gap from the optimal $O(\log T)$ regret possible. This inspired work on \textit{Maler} by \cite{maler_guanghui}. The main idea was to run multiple learning algorithms for each of the three function classes in parallel. The master algorithm continued to track the best one on the fly. This spawned three times the number of experts as Metagrad, but enjoyed optimal regret bound on all three function classes. We borrow inspirations the learning algorithm employed by Maler on the strongly convex case and extend it to handle sparse gradients.

\section{Setup}
\subsection{Notation}
We denote $C\subseteq \mathbb{R}^d$ to be our convex decision set. We let $\x_t \in C$ be the learner's decision at time step $t$. We denote our loss function at time step $t$ to be $f_t(\cdot) : C \rightarrow \mathbb{R}$. We let $\g_t$ be the gradient of $f_t(\x_t)$. We use $\lVert \cdot \rVert$ to denote the 2-norm, and $\langle \cdot , \cdot \rangle$ to denote the Euclidean inner product. We use boldface letters to denote vectors.
\subsection{Assumptions and Definitions}
We include two main assumptions akin to those from previous studies \cite{maler_guanghui}. \newline \newline 
\textbf{Assumption 1: } The gradients of all loss functions $f_1(\cdot) \dots f_T(\cdot)$ are bounded i.e. $\forall t \in [T], \forall \x \in C$ $ \lVert \nabla f_t(\x)\rVert \leq G$ for some $G \in \mathbb{R}$.\newline
\textbf{Assumption 2: } The diameter of the decision set $C$ is bounded i.e. $\forall \x_1, \x_2 \in C$, $\lVert \x_1 - \x_2 \rVert \leq D$ for some $D \in \mathbb{R}$. \newline  
\textbf{Definition 1: Strong Convexity} A function $f: C \rightarrow \mathbb{R}$ is $\lambda$-strongly convex if $\forall \x_1,\x_2 \in C$, 
$$ f(\x_1) \geq f(\x_2) + \langle \nabla f(\x_2), \x_1 - \x_2 \rangle + \frac{\lambda}{2} \lVert \x_1 - \x_2 \rVert^2$$
\textbf{Assumption 3: } All loss functions $f_1(\cdot)\dots f_T(\cdot)$ are $\lambda$-strongly convex. \newline
\textbf{Definition 2: A-weighted norm} 
Let $A \in \mathbb{R}^{d \times d}$ and let $\x \in \mathbb{R}^d$. We define the $A$-weighted norm to be the matrix norm $\lVert \x \rVert_A^2 = \x^\top A \x = \sum_{i,j=1}^d A_{ij}\x_i\x_j$ \cite{SC_adagrad}. \newline
\textbf{Definition 3: Weighted Projection}
We define a weighted projection function onto the convex set $C$ with respect to the $A$-weighted norm to be 
$$\Pi_C^A(\x_1) = \arg\min_{\x \in C} \lVert \x_1-\x  \rVert^2_A$$

\section{Algorithm}
The algorithm draws inspiration from the strongly convex experts in Maler. It similarly follows a two hierarchical structure. At the lower level, we have a set of experts, each with their own learning rates $\eta$. Because we are only dealing with the strongly convex case, all of them use the same learning algorithm (Algorithm 2). At the higher level, we have a meta algorithm which tracks the best one using a tilted exponential weighted average \cite{metagrad}.
% TODO: We need to talk about the different learning rates assigned to each slave, and how they depend on the $\mu$ (where mu is the strong convexity parameter).
\begin{algorithm}
\label{algo:meta_algorithm}
\caption{Meta Algorithm}
\KwIn{Grid of learning rates $\eta_1,\eta_2,\dots,\eta_k$ and weights $\pi_1^{\eta_1},\pi_1^{\eta_2},\dots ,\pi_1^{\eta_k}$}{}
Get prediction $\x_t^\eta$ from Algorithm 2 \newline
Play $\x_t = \frac{\sum_{\eta} \pi_t^\eta \eta \x_t^\eta }{\sum_{\eta} \pi_t^\eta \eta }$ \newline
Observe gradient $\g_t$ and send $\x_t, \g_t$ to all experts \newline 
For all $\eta$, update weights according to 
$$\pi_{t+1}^\eta = \frac{\pi_t^\eta \exp{\left (-s_t^\eta (\x_t^\eta )\right )}}{\sum_{\eta } \pi_t^\eta \exp{\left (-s_t^\eta (\x_t^\eta \right ))}}$$
\end{algorithm}

\noindent We define the surrogate loss function to be 
\begin{equation}
s_t^\eta(\x)=-\eta (\x_t -\x)^\top \g_t + \eta^2 G^2 \lVert \x_t - \x \rVert^2 
\end{equation}
Note, by taking the second derivative of the surrogate loss function, we see that it is $\eta^2G^2$ strongly convex.\\

\begin{algorithm}
\label{algo:expert_algorithm}
\caption{SC-Adagrad expert algorithm}
\KwIn{$\velocity_0=\mathbf{0} \in \mathbb{R}^d$,$ \eta \in \mathbb{R}$, $\x_1^\eta = 0$, $\delta > 0$}{}
\For {$t = 1\dots T$} 
{ 
Send $\x_t^\eta$ to Algorithm 1 \newline
Receive $\x_t, \g_t$ from Algorithm 1 \newline
$\nabla s_t^\eta (\x_t^\eta) = \eta \g_t + 2\eta^2 G^2(\x_t^\eta-\x_t)$ \newline
$\velocity_t = \velocity_{t-1} + \nabla \stxt \odot \nabla \stxt$ \newline
$A_t = \text{diag}(\velocity_t) + \delta I$ \newline 
$\x_{t+1}^\eta = \Pi_C^{A_t}\left(\x_t^\eta - \alpha A^{-1}_t \nabla s_t^\eta (\x_t^\eta) \right )$, where $\alpha = \frac{1}{4\eta^2}$
}

\end{algorithm}
\noindent \textbf{Experts:} Each expert uses an adaptive algorithm with an update rule derived from the Adagrad for strongly convex function \cite{SC_adagrad}. However, in contrast to the original Adagrad, this adaptation is on the gradients of the surrogate losses, rather than the $\g_t$ incurred by a single learner. We maintain $\lceil\frac{1}{2}\log_2 T\rceil$ experts. For $i = 0,1,2,...\frac{1}{2}\log_2 T$, each expert is configured with:
$$ \eta_i= \frac{2^{-i}}{5GD}$$
and
$$\pi_1^{\eta_i} = \frac{c}{3(i+1)(i+2)}$$
These values for $\eta$, $\pi$ and $\alpha$ were carefully chosen to satisfy certain criteria which will help in proving regret bounds.

\section{Regret Analysis}
\subsection{Preliminaries}
% \begin{lemma}
\subsubsection*{Lemma 1}
\label{scadagrad_bound}
 [Lemma 3.2 from \cite{SC_adagrad}] Let assumptions \textbf{A1} and \textbf{A2} hold, then for all $T\geq 1$ and $A_t, \delta_t$ defined from algorithm 2 and $\nabla s_t = \stxt$, we have:
$$\addt \langle \nabla s_t, A^{-1}_t \nabla s_t  \rangle \leq \sum_{i=1}^d \log \Bigg( \frac{||\nabla s_{1:T,i}||^2 + \delta}{\delta}\Bigg)$$
% \end{lemma}

% \begin{lemma}
\subsubsection*{Lemma 2}
\label{norm_subscript_bounds} $\forall t$, let $A_t$ be positive semi-definite diagonal matrices as which follows the update rules in Algorithm 2. Similarly, let $\alpha \geq \frac{1}{4\eta^2}$ for some constant $\eta$ and let $\mu = 2\eta^2 G^2$. Then the following holds:
\begin{align*}
    A_t - A_{t-1} - 2\alpha \text{diag}(\mu) \preceq 0 
\end{align*}
and
\begin{align*}
    A_1 - 2\alpha \text{diag}(\mu) \preceq \delta
\end{align*}
% \end{lemma}
\textbf{Proof:}
In our update rule we have:
\begin{align*}
    A_t &= \text{diag}(v_{t}) + \text{diag}(\delta)\\
    &= v_{t-1} + \nabla \stxt \bullet \nabla \stxt + \delta \\
    &= A_{t-1} + \nabla \stxt \bullet \nabla \stxt \\
    &\leq A_{t-1} + {(G^s)}^2
\end{align*}
\noindent where, $\nabla \stxt \bullet \nabla \stxt $ has an upper bound of $(G^s)^2$.\\
This can be rearranged as:
\begin{align*}
    A_t - A_{t-1} &\leq {(G^s)}^2\\
    \Rightarrow A_t - A_{t-1} - 2\alpha \text{diag}(\mu) &\leq {(G^s)}^2 - 2\alpha \text{diag}(\mu)\\
    &\leq {(G^s)}^2 - 2 \left(\frac{1}{4\eta^2 }\right)(2 \eta^2 {(G^s)}^2)\\
    &\leq {(G^s)}^2 - {(G^s)}^2 = 0
\end{align*}
This proves our the first inequality. For the second one, since we have $\velocity_0 =0$ in our algorithm, we observe that $A_1 = \stxt+\stxt + \text{diag}(\delta)$. With the same setting of $\alpha$, we have:
\begin{align*}
    A_1 &\leq {(G^s)}^2 +\delta \\
    A_1 - 2\alpha \text{diag}(\mu) &\leq {(G^s)}^2  - 2\alpha \text{diag}(\mu) +\delta \\
    &\leq \delta
\end{align*}
This proves the second inequality.
% Where in the last step we added a term on both sides. Now if we have $\alpha \geq \frac{{G^s}^2}{2 diag(\mu)}$, then the following can be simplified:

\subsection{Meta Regret}
The regret of the meta algorithm is defined as the difference of the surrogate losses of the meta learner and the surrogate losses of a given expert. It is given by:
\begin{align*}
    R_T^{Meta} = \addt s_t(\x_t) - \addt \stxt \tag{5.2.1}
\end{align*}

% \begin{lemma}
\subsubsection*{Lemma 3}
\label{meta_regret_label}
[Lemma 1 in \cite{maler_guanghui}] For every grid point $\eta$ we have:
$$\addt s_t(\x_t) - \addt \stxt \leq 2\ln \bigg(\sqrt{3} \left (\frac{1}{2}\log_2 T + 3\right )\bigg)$$
% \end{lemma}

The proof is attached in the appendix.
\subsection{Expert Regret}

% Mention expert regret using results from SC adagrad. PROVE IT HERE since this is a major result
The regret of the expert algorithm is defined as the difference of the surrogate losses of the expert and the surrogate losses of the best decision in hindsight. It is given by:
\begin{align*}
    R_T^{Expert} = \addt \stxt - \addt s_t^\eta(\x^*) \tag{5.3.1}
\end{align*}

Let us denote $\nabla s_t = \grads$\\
% \begin{lemma}
\subsubsection*{Lemma 4}
\label{expert_prelim}
\begin{align*}
    (\x_t^\eta - \x^*)^\top \nabla s_t &\leq \frac{||\xte - \x^*||^2_{A_t} - ||\x_{t+1}^\eta - \x^*||^2_{A_t}}{2\alpha} + \frac{\alpha}{2} \langle \nabla s_t, A_t^{-1}\nabla s_t \rangle
\end{align*}
% \end{lemma}
\textbf{Proof : } \\
From the update rule, we have :
\begin{align*}
    ||\x_{t+1}^\eta - \x^*||_{A_t} &= \left \|\Pi_C^{A_t}(\xte - \alpha A_t^{-1} \nabla s_t - \x^*)\right \|^2_{A_t} \\
    &\leq \left \|\xte - \alpha A_t^{-1} \nabla s_t - \x^*\right \|^2_{A_t} \\
    &= ||\xte -\x^*||^2_{A_t} + \alpha^2||A_t^{-1} \nabla s_t||^2_{A_t} - 2 \alpha (\xte -\x^*)^\top A_t (A_t^{-1}\nabla s_t) \\
    \Rightarrow 2 \alpha (\x_t^\eta - \x^*)^\top \nabla s_t &\leq ||\xte -\x^*||^2_{A_t} - ||\x_{t+1}^\eta - \x^*||_{A_t} + \alpha^2||A_t^{-1} \nabla s_t||^2_{A_t} \\
    \Rightarrow (\x_t^\eta - \x^*)^\top \nabla s_t &\leq \frac{||\xte -\x^*||^2_{A_t} - ||\x_{t+1}^\eta - \x^*||_{A_t}}{2\alpha} + \frac{\alpha}{2}\langle \nabla s_t, A_t^{-1}\nabla s_t \rangle
\end{align*}

% \begin{lemma}
\subsubsection*{Lemma 5}
\label{expert_regret_lemma}
The expert regret is given by:
\begin{align*}
    R_T^{Expert} = \addt \stxt - \stxs \leq \frac{D^2 d\delta}{2\alpha} + \frac{\alpha}{2} \sum_{i=1}^d \log \Bigg( \frac{||\nabla s_{1:T,i}||^2 + \delta}{\delta}\Bigg)
\end{align*}

% \end{lemma}
\noindent \textbf{Proof:}
We can start from the definition of strong convexity
\begin{align*}
    \addt \stxt - \stxs &\leq \addt \langle \grads, \x_t-\x^*\rangle - ||\xte - \x^*||^2_{\text{diag}(\mu)} \\
    &\leq \addt \frac{||\xte -\x^*||^2_{A_t} - ||\x_{t+1}^\eta - \x^*||^2_{A_t}}{2\alpha} + \frac{\alpha}{2}\langle \nabla s_t, A_t^{-1}\nabla s_t \rangle - ||\xte - \x^*||^2_{\text{diag}(\mu)} \tag{Using Lemma \ref{expert_prelim}}\\
    &= \frac{||\x_1^\eta - \x^*||^2_{A_1}}{2\alpha} + \sum_{t=2}^T \frac{||\xte -\x^*||^2_{A_t} - ||\xte - \x^*||^2_{A_{t-1}}}{2\alpha} - \frac{||\x_{T+1}^\eta - \x^*||^2_{A_T}}{2\alpha}  \\
    &+ \addt \frac{\alpha}{2}\langle \nabla s_t, A_t^{-1}\nabla s_t \rangle - \addt ||\xte - \x^*||^2_{\text{diag}(\mu)} \tag{Separating the first and last terms}\\
    &\leq \frac{||x_1^\eta - \x^*||^2_{A_1 - 2\alpha \text{diag}(\mu)}}{2\alpha} + \sum_{t=2}^T \frac{||\xte -\x^*||^2_{A_t - A_{t-1} - 2\alpha \text{diag}(\mu)}}{2\alpha}\\ 
    &+ \frac{\alpha}{2}\langle \nabla s_t, A_t^{-1}\nabla s_t \rangle\\
    &\leq \frac{||\x_1^\eta - \x^*||^2_{\delta}}{2\alpha} +\sum_{t=2}^T \frac{||\xte -\x^*||^2_0}{2\alpha}+ \frac{\alpha}{2}\langle \nabla s_t, A_t^{-1}\nabla s_t \rangle  \tag{Using Lemma \ref{norm_subscript_bounds}}\\
    &\leq \frac{||\x_1^\eta - \x^*||^2_{\delta}}{2\alpha} + \frac{\alpha}{2}\langle \nabla s_t, A_t^{-1}\nabla s_t \rangle  \tag{because second term $\leq$ 0}\\
    &\leq \frac{D^2 d\delta}{2\alpha} + \frac{\alpha}{2}\langle \nabla s_t, A_t^{-1}\nabla s_t \rangle  \tag{Using A1}\\
    &\leq \frac{D^2 d\delta}{2\alpha} + \frac{\alpha}{2} \sum_{i=1}^d \log \Bigg( \frac{||\nabla s_{1:T,i}||^2 + \delta}{\delta}\Bigg) \tag{Using Lemma \ref{scadagrad_bound} }\\
\end{align*}

\subsection{Total Regret}
The total regret is given by 
\begin{align*}
    R_T &= \addt f_t(\x_t) - f_t(\x^*) \\
    &\leq \addt \g_t^\top (\x_t - \x^*) - \frac{\mu}{2}||\x_t - \x^*||^2 \tag{from definition of strongly convex functions}\\
    &= \addt \frac{-s_t^\eta(\x^*) + \eta^2G^2||\x_t-\x^*||}{\eta} - \frac{\mu}{2}||\x_t - \x^*||^2 \tag{from the definition of surrogate loss function}\\
    &\leq \addt \frac{s_t^\eta(\x_t)-s_t^\eta(\x^*)}{\eta} + \eta \addt G^2||\x_t-\x^*|| - \frac{\mu}{2}\addt ||\x_t - \x^*||^2 \\
    &= \addt \frac{s_t^\eta(\x_t)-\stxt}{\eta} + \addt \frac{\stxt-s_t^\eta(\x^*)}{\eta} + \eta \addt G^2||\x_t-\x^*|| - \frac{\mu}{2}\addt ||\x_t - \x^*||^2 \\
    &\leq \frac{2}{\eta} \log\left(\sqrt{3}\left(\frac{1}{2} \log(T) + 3\right)\right) + \frac{1}{\eta} \frac{D^2 d\delta}{2\alpha} + \frac{\alpha}{2} \sum_{i=1}^d \log \Bigg( \frac{||\nabla s_{1:T,i}||^2 + \delta}{\delta}\Bigg) \tag{Using lemma \ref{meta_regret_label}, \ref{expert_regret_lemma}}\\
    & + \eta V_T^S - \frac{\mu}{2}\addt ||\x_t - \x^*||^2 \tag{5.4.1}\\
\end{align*}
where, $V_T^S = \addt G^2||x_t-x^*||^2$.

\noindent Let us denote 
\begin{align*}
    E_T &= \frac{D^2 d\delta}{2\alpha} + \frac{\alpha}{2} \sum_{i=1}^d \log \Bigg( \frac{||\nabla s_{1:T,i}||^2 + \delta}{\delta}\Bigg) = O\left(d \log\left(\frac{T}{\delta}\right)\right) \tag{5.4.2} \\
    A_T &= 2\log\left(\sqrt{3}\left(\frac{1}{2} \log(T) + 3\right)\right) + E_T \tag{5.4.2}
\end{align*}

\noindent Using (5.4.2) in (5.4.1), we have
\begin{align*}
    R_T &\leq \frac{A_T}{\eta} + \eta V_T^S - \frac{\mu}{2}\addt ||\x_t - \x^*||^2 \tag{5.4.3}
\end{align*}

\noindent The optimal $\hat{\eta}$ to minimize $A_T + \eta V_T^S$ is (\cite{maler_guanghui})
\begin{align*}
    \hat{\eta} = \sqrt{\frac{A_T}{V_T^S}} \geq \frac{1}{5GD} \tag{5.4.4}
\end{align*}

\noindent If $\hat{\eta} \leq \frac{1}{5GD}$, then by construction there exists a grid point $\eta \in [\frac{\hat{\eta}}{2}, \hat{\eta}]$. Thus.
\begin{align*}
    R_T &\leq \eta V_T^S + \frac{A_T}{\eta} - \frac{\mu}{2}\addt ||\x_t - \x^*||^2\\
    &\leq \hat{\eta} V_T^S + \frac{2A_T}{\hat{\eta}} - \frac{\mu}{2}\addt ||\x_t - \x^*||^2\\
    &= 3\sqrt{V_T^S A_T} - \frac{\mu}{2}\addt ||\x_t - \x^*||^2 \tag{5.4.5}\\
\end{align*}

\noindent If $\hat{\eta} \geq \frac{1}{5GD}$, by (5.4.4), we have
\begin{align*}
    V_T^S \leq 25G^2D^2A_T \tag{5.4.6}
\end{align*}

\noindent Using (5.4.4), (5.4.6) and $\hat{\eta} = \frac{1}{5GD}$ in (5.4.3), we have
\begin{align*}
    R_T 
    &\leq 2\sqrt{A_T V_T^S} - \frac{\mu}{2}\addt ||\x_t - \x^*||^2\\
    &\leq 10 GDA_T - \frac{\mu}{2}\addt ||\x_t - \x^*||^2\tag{5.4.7}\\
\end{align*}

\noindent Hence, combining (5.4.5) and (5.4.6), we have 
\begin{align*}
    R_T &\leq 3\sqrt{V_T^S A_T} + 10GDA_T - \frac{\mu}{2}\addt ||\x_t - \x^*||^2 \tag{5.4.8}\\
\end{align*}

\noindent Let $x = V_T^S$ and $y = A_T$. For $x,y,\gamma \geq 0$
\begin{align*}
    \sqrt{xy} &\leq \frac{\gamma}{2}x + \frac{1}{2\gamma}y \\
    \Rightarrow 3\sqrt{V_T^SA_T} &\leq \frac{3\gamma}{2}V_T^S + \frac{3}{2\gamma}A_T \tag{5.4.9}
\end{align*}

\noindent Using (5.4.9) in (5.4.8), we have

\begin{align*}
    \noindent R_T &\leq \frac{3\gamma}{2}V_T^S + \frac{3}{2\gamma}A_T + 10GDA_T - \frac{\mu}{2}\addt ||x_t - x^*||^2 \\
    &= \frac{3\gamma}{2}V_T^S + \frac{3}{2\gamma}A_T + 10GDA_T - \frac{\mu}{2G^2}V_T^S \\
    &= \left(\frac{3\gamma}{2}- \frac{\mu}{2G^2}\right)V_T^S + \left(\frac{3}{2\gamma} + 10GD\right)A_T \\
    \text{Setting } \gamma &= \frac{\mu}{3G^2}\text{, we have} \\
    R_T &= \left(\frac{G^2}{\mu} + 10GD\right)A_T \\
    &= \left(\frac{G^2}{\mu} + 10GD\right)\left(2\log\left(\sqrt{3}\left(\frac{1}{2} \log(T) + 3\right)\right) + \frac{D^2 d\delta}{2\alpha} + \frac{\alpha}{2} \sum_{i=1}^d \log \Bigg( \frac{||\nabla s_{1:T,i}||^2 + \delta}{\delta}\Bigg) \right)\\
    % &= O\left(d\log\left(\frac{T}{\delta}\right)\right)\\
\end{align*}

\textbf{Remark :} In \cite{SC_adagrad} the authors claim that
\begin{align*}
    \sum_{i=1}^d \log \Bigg( \frac{||\nabla s_{1:T,i}||^2 + \delta_{T,i}}{\delta_{1,i}}\Bigg) = O(\log T)
\end{align*}

\noindent And hence, the overall regret is $O(\log T)$ for strongly-convex functions. Even with the monotonic-decay factor, the term 
\begin{align*}
    \frac{||\nabla s_{1:T,i}||^2}{\delta_{1,i}} \leq \frac{TG^2}{\delta_{1,i}}
\end{align*}
\noindent Hence, the regret term is bound by $O(d\log(T))$ rather than $O(\log(T))$. This result hasn't been explained in the paper which seems like a weakness on the results. Hence, we take state the final regret bound as $O(d \log(T))$.

% BY HASSAN: parts about showing the paragraph in SC adagrad paper

\section{Conclusion and Future Work}
By taking a Metagrad like approach to Adagrad, we can have created a parameter free version of Strongly Convex Adagrad. This enjoys $O(d\log T)$ regret bound in strongly convex functions. If every expert employs SAdam \cite{Wang2020SAdam:}, the regret bound can be improved to $O(log T)$ instead. This work can be extended to make parameter free versions of other algorithms such as RMSProp or Adam.

%Bibliography
\bibliographystyle{unsrt}  
\bibliography{references}  

\appendix

\section{Proof of Meta Regret}

This proof follows closely from the proof of the meta regret bound in \cite{maler_guanghui}. We define a potential function 
\begin{align*}
\Phi(T) = \sum_{\eta } \pi_1^\eta \exp \left (-\sum_{t=1}^T s_t^\eta (\x_t^\eta)\right )
\end{align*}
We have 
\begin{align*}
\Phi(T+1)-\Phi(T) &= \sum_{\eta} \pi_1^\eta \exp\left (-\sum_{t=1}^T s_t^\eta (\x_t^\eta) \right ) \exp \left (-s_{T+1}^\eta \left (\x_{T+1}^\eta \right ) - 1 \right ) \\
&\leq \sum_{\eta} \pi_1^\eta \exp\left (-\sum_{t=1}^T s_t^\eta (\x_t^\eta) \right ) \eta (\x_{T+1} - \x_{T+1}^\eta)^\top \g_t \\
&= (\avector_T\x_{T+1}-\bvector_T)^\top \g_t
\end{align*}
where 
\begin{align*}
\avector_T &= \sum_{\eta} \pi_1^\eta \exp\left (-\sum_{t=1}^T s_t^\eta (\x_t^\eta) \right ) \eta \\
\bvector_T &= \sum_{\eta} \pi_1^\eta \exp\left (-\sum_{t=1}^T s_t^\eta (\x_t^\eta) \right ) \eta \x_{T+1}^\eta 
\end{align*}
By definition, 
\begin{align*}
\x_{T+1} &= \frac{\sum_{\eta} \pi_{T+1}^\eta \eta \x_{T+1}^\eta }{\sum_{\eta} \pi_{T+1}^\eta \eta } \\
&= \frac{\sum_{\eta} \pi_1^\eta \exp\left (-\sum_{t=1}^{T+1} s_t^\eta(\x_t^\eta ) \right )\eta \x_{T}^\eta}{\sum_{\eta } \pi_1^\eta \exp(-\sum_{t=1}^{T} s_t^\eta (\x_t^\eta))\eta }\\
&= \frac{\bvector_T}{\avector_T}
\end{align*}
This implies that $\Phi(T+1)-\Phi(T) \leq 0$ and $1=\Phi(0)\geq \Phi(1) \geq \dots \Phi(T) \geq 0$. Then for any $\eta$
\begin{align*}
\ln\left ( \pi_1^\eta \exp \left (-\sum_{t=1}^T s_t^\eta (\x_t^\eta)\right )\right ) &\leq 0 \\
\implies \ln\left(\frac{1}{\pi_1^\eta}\right ) +\sum_{t=1}^T s_t^\eta (\x_t^\eta) \geq 0
\end{align*}
By definition, $\pi_1^{\eta_i} = \frac{c}{3(i+1)(i+2)}$, where $c=1+\frac{1}{1+k}$, $i\in \{0,\dots, k\}$ and $k= \left \lceil \frac{1}{2}\log_2(T) \right \rceil $. Therefore, 
\begin{align*}
\ln\left(\frac{1}{\pi_1^\eta}\right ) &= \ln\left (\frac{3(i+1)(i+2)}{c} \right) \\
&\leq \ln\left( 3(1+k)(2+k) \right) \\
&\leq \ln\left (\left ( \sqrt{3}\left (3+\frac{1}{2}\log_2(T)\right )\right )^2\right ) \\
&= 2 \ln \left ( \sqrt{3}\left (3+\frac{1}{2}\log_2(T)\right )\right )
\end{align*}
\end{document}